\documentclass{article} 
\usepackage{iclr2027_conference,times}

\usepackage{amsmath,amsfonts,bm}

\def\eqref#1{equation~\ref{#1}}

\def\1{\bm{1}}

\DeclareMathAlphabet{\mathsfit}{\encodingdefault}{\sfdefault}{m}{sl}
\SetMathAlphabet{\mathsfit}{bold}{\encodingdefault}{\sfdefault}{bx}{n}

\usepackage[pagebackref,breaklinks,colorlinks,allcolors=gray]{hyperref}
\usepackage{url}
\usepackage{booktabs}
\usepackage{multirow}
\usepackage{makecell}
\usepackage{graphicx}
\usepackage{caption}
\usepackage{algorithm}
\usepackage{algpseudocode}
\usepackage{colortbl}
\def\methodname{WinARDet}

\title{\methodname{}: Compressed Window Auto-regression for Long-Horizon Vision-Language Grounding}

\author{
Chenxu Wang$^{1}$, Hangyi Kuang$^{1}$, Bowen Yin$^{1}$, Yuxuan Li$^{1}$, Quansheng Zeng$^{1}$ \\
Yang Zhang$^{1}$, Xu Guo$^{2}$, Xiang Li$^{1}$, Qibin Hou$^{1}$\thanks{Corresponding author. Email: \texttt{houqb@nankai.edu.cn}.} \\[0.5em]
$^{1}$Nankai University \qquad $^{2}$Beijing Normal University
}

\iclrfinalcopy 
\begin{document}

\maketitle

\begin{abstract}
    MLLM-based detectors formulate vision-language grounding as autoregressive coordinate generation, yet in dense scenes, they often collapse into a box repetition loop, repeatedly emitting identical boxes until the decoding budget is exhausted. We associate this failure with a historical-answer attention takeover: Under full-history autoregression, repeated coordinate tokens in the growing answer history attract disproportionate attention, crowding out the attention sinks needed to attend to new visual evidence. 
    To break this loop, we propose \textbf{Compressed Window Auto-Regression} (WinAR). This new mechanism partitions a long answer sequence into consecutive generation windows and restricts full token-level autoregression to the current window only while compressing each previous window into a fixed set of latent tokens that condition subsequent decoding. This design reduces the repeated exposure to raw historical coordinates that triggers the attention takeover, yet preserves cross-window dependencies essential for coherent multi-object generation. 
    On RSDense, which contains 200-2,000 objects per image, our model \methodname{} improves RexOmni F1-score (IoU=0.5) from 32.16 to 39.36 over the baseline, while reducing the redundancy rate from 35.48\% to 13.00\%. 
    \methodname{} also generalizes across decoding strategies and output formats, including parallel box decoding and polygon generation, providing a practical paradigm for long-horizon vision-language grounding across scene densities. 
    Code: \url{https://github.com/facias914/WinARDet}

\end{abstract}

\section{Introduction}
\label{sec:1}

Modern MLLM-based detectors, such as RexOmni~\citep{jiang2026detect} and LocateAnything~\citep{wang2026locateanything}, formulate object detection as autoregressive next-token prediction, enabling flexible vision-language grounding under complex instructions beyond the fixed prediction interfaces of conventional detectors~\citep{liu2024grounding, cheng2024yolo, wang2026slip}. However, extending this autoregressive formulation to dense visual scenes introduces a fundamental long-horizon grounding challenge~\citep{xia2018dota, goldman2019precise}. 
When hundreds or even thousands of objects must be generated within a single response, the model must continuously discover previously uncovered objects while keeping track of those already localized. Since autoregressive inference operates under a finite and costly decoding budget, every redundant prediction consumes generation steps that could otherwise be used to recover new objects. Sustaining new-object discovery throughout a long decoding horizon is therefore critical to both detection accuracy and inference efficiency. 

\begin{figure}[h]
   \centering
   \includegraphics[width=\linewidth]{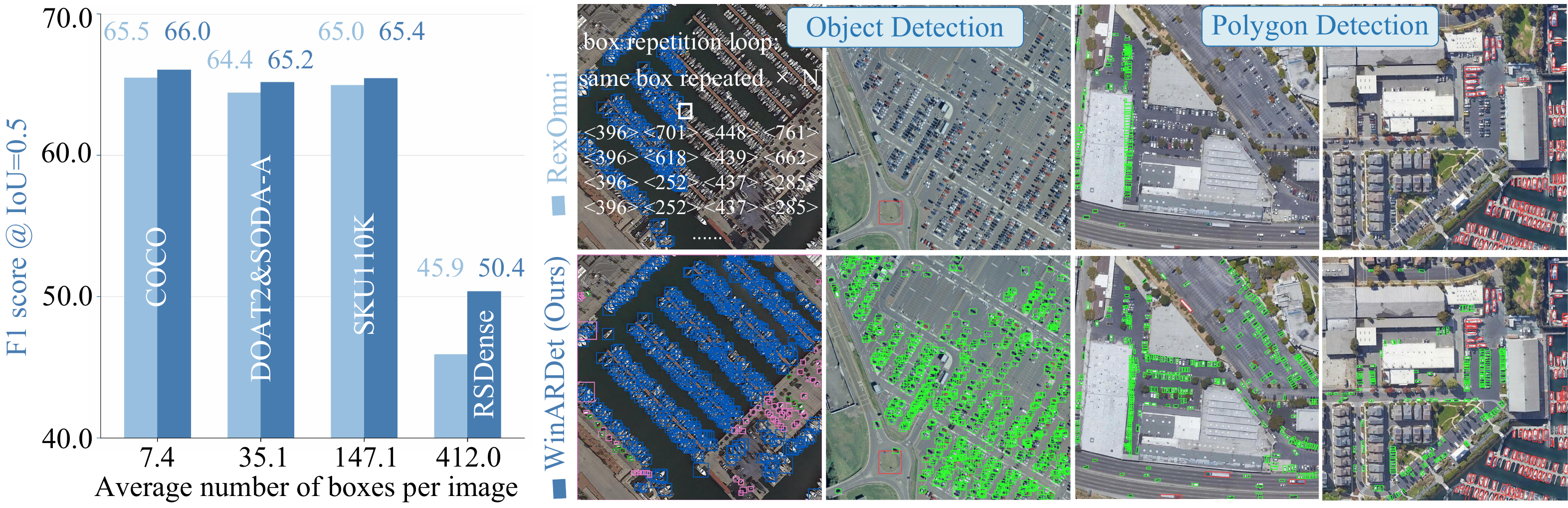}
   \caption{
    \methodname{} overview. 
      \textbf{Grounding Performance across Densities(Left):} WinARDet outperforms the baseline across benchmarks of varying scene densities(~\cite{lin2014microsoft, xia2018dota, cheng2023towards, goldman2019precise}), with increasingly pronounced gains in denser scenes. 
      \textbf{Breaking Box Repetition Loop(Middle):} WinARDet breaks the box repetition loop, suppressing repeated predictions and recovering substantially more distinct objects. 
      \textbf{Beyond Bounding Boxes(Right):} WinARDet generalizes to polygon detection, extending beyond bounding-box. 
   }
   \label{fig1}
   \vspace{-10pt}
\end{figure}

\begin{figure}[t]
   \centering
   \includegraphics[width=\linewidth]{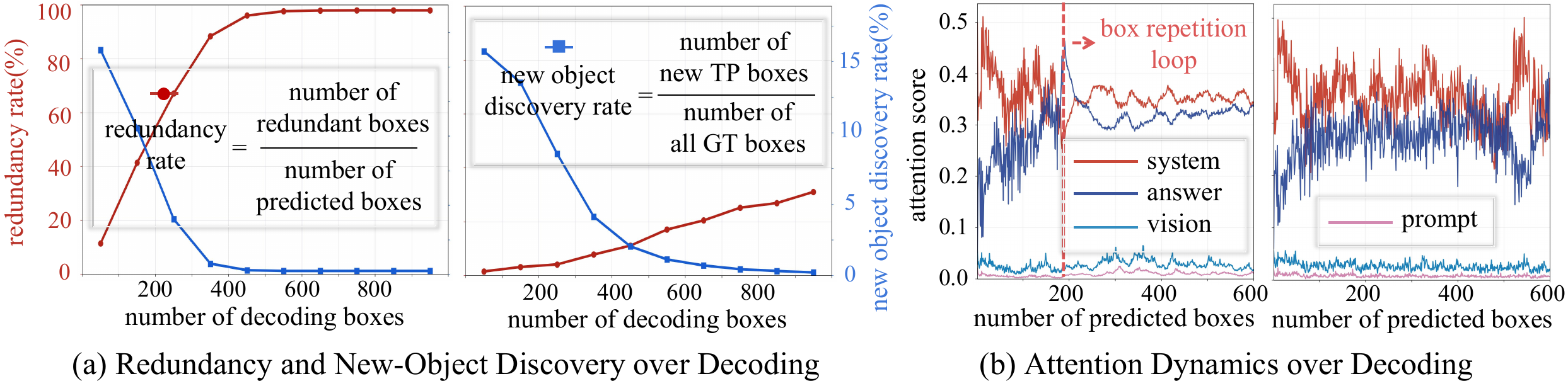}
   \caption{
      \textbf{(a)} Metrics are averaged over intervals of 100 predicted boxes on RSDense(Section~\ref{sec:exp}). 
      FHAR(left) rapidly accumulates redundant predictions as new-object discovery collapses, whereas WinAR(right) suppresses redundancy and sustains the discovery of new objects. 
      \textbf{(b)} The onset of the box repetition loop coincides with historical-answer attention takeover and reduced system-token attention in FHAR(left), while WinAR(right) maintains a more stable attention allocation. 
   }
   \label{fig2}
\end{figure}

Our empirical investigation reveals that MLLM-based detectors under the Full-History Auto-Regression (FHAR) paradigm, where each prediction conditions on the complete history of generated answer tokens, progressively lose their ability to sustain new-object discovery over extended decoding horizons.
In dense configurations, we observe that a supervised fine-tuned RexOmni model~\citep{jiang2026detect} under the FHAR paradigm initially yields valid detections, yet eventually deteriorates into a pathological box repetition loop, wherein identical or nearly identical bounding boxes are emitted repeatedly (Fig.~\ref{fig1}(middle, top)). 
This decoding collapse is quantitatively characterized in Fig.~\ref{fig2}(a): Over the initial 400 box predictions, the redundancy rate escalates from approximately 10\% to near 100\%, while the rate of discovering novel objects concomitantly plummets from roughly 16\% to nearly zero. 
Consequently, a substantial portion of the remaining decoding budget is expended on redundant box predictions rather than on unaccounted objects, causing recall to saturate while inference latency continues to accumulate. 
Ultimately, this degenerative process substantially undermines both the detection accuracy and the overall inference efficiency. 

Our attention analysis links this degeneration to a progressive historical-answer attention takeover. 
Specifically, we partition the conditioning context into system, visual, prompt, and historical-answer regions, and track the softmax-normalized attention mass assigned to each region by coordinate-token queries throughout decoding. 
As shown in Fig.~\ref{fig2}(b), attention to the historical answer grows steadily with the decoding horizon and rises sharply near the onset of the repetition loop, accompanied by a pronounced reduction in attention to the initial system tokens. 
This redistribution is particularly revealing because prior studies~\citep{xiao2024efficient, qiu2026gated} have shown that a small set of initial system tokens often serves as attention sinks, absorbing excess attention during autoregressive decoding. 
In dense grounding, however, the expanding raw answer history progressively displaces this sink allocation with explicit and increasingly repetitive coordinate patterns. 
Once these raw coordinates dominate the conditioning context, the decoder becomes increasingly biased toward previously generated locations. Each duplicated box is then appended back to the history, further strengthening this bias and ultimately forming a self-reinforcing box repetition loop. 

To break this loop, we propose \textbf{Compressed Window Auto-Regression} (\textbf{WinAR}), an alternative autoregressive paradigm for long-horizon structured generation. 
WinAR follows a simple principle: \emph{remember compactly and generate locally}. 
In contrast to FHAR, which retains every previous coordinate token, WinAR preserves exact token-level history only within the current window and represents each previous completed window with a fixed number of latent tokens by a Window History Compressor. 
During training, preceding windows appear only through their compact latent tokens, and cross-entropy supervision is applied to the current window. 
During inference, each completed window is compressed before generation proceeds to the next. 
WinAR therefore changes how autoregressive context is carried across a long answer: local generation remains exact, while cross-window dependencies pass through a compact history. 
By limiting repeated exposure to raw coordinate patterns without eliminating historical conditioning, WinAR mitigates historical-answer attention takeover and preserves the decoding budget for discovering new objects. 

Experiments show that the benefit of WinAR becomes increasingly pronounced as scene density grows~\citep{xia2018dota, cheng2023towards}. 
On RSDense benchmark, which contains 200--2,000 objects per image, integrating WinAR into RexOmni reduces the mean prediction redundancy rate from 35.48\% to 13.00\% and improves F1@IoU$=0.5$ from 32.16 to 39.36. 
It also generalizes to the parallel box decoding framework of LocateAnything~\citep{wang2026locateanything} and improves F1 from 26.15 to 27.54 and reduces mean redundancy from 14.89\% to 2.13\% on RSDense. 
WinAR also extends to structured outputs such as polygons and substantially reduces peak training memory. 
When training LocateAnything, WinAR reduces peak GPU memory consumption by 34.1\%, enabling training on 8 NVIDIA A40 GPUs(48~GB) without MagiAttention~\citep{zewei2025magiattention}. 
These results establish WinAR as a general and practical paradigm for accurate and efficient vision-language grounding across varying scene densities. 
Our contributions are threefold:
\begin{itemize}
    \item We identify a long-horizon failure mode in MLLM-based detectors, where Full-History Auto-Regression progressively triggers historical-answer attention takeover and a self-reinforcing box repetition loop. 
    \item We propose \textbf{Compressed Window Auto-Regression (WinAR)}, which preserves exact autoregression within local windows while compressing completed windows into compact latent tokens, mitigating repetition without discarding cross-window dependencies. 
    \item Extensive experiments across scene densities, decoding strategies, and structured output formats demonstrate that WinAR consistently improves detection accuracy, substantially reduces prediction redundancy, and lowers training memory for long-sequence grounding. 
\end{itemize}

\section{Related Work}
\label{sec:2}

\textbf{Regression-based Object Detection}~\citep{ren2016faster, cai2018cascade, carion2020end, zhu2020deformable, redmon2016you, Dai_2021_CVPR} is built upon a discriminative classification and regression paradigm. 
They typically extract multi-scale feature maps with a backbone~\citep{dosovitskiy2020image, simeoni2025dinov3} and a feature pyramid network~\citep{lin2017feature}, instantiate candidate proposals in the form of anchors~\citep{ren2016faster}, points~\citep{tian2019fcos}, or learnable object queries~\citep{carion2020end}, assign foreground and background samples via label assignment strategy~\citep{zhang2020bridging, wang2025multi}, optimize semantic recognition and geometric localization with classification~\citep{lin2017focal} and regression losses~\citep{li2020generalized}, respectively. 
Motivated by large-scale contrastive vision-language pre-training such as CLIP~\citep{radford2021learning}, open-vocabulary detectors~\citep{li2022grounded, liu2024grounding} further align category names or textual prompts with regional visual features, extending detection beyond fixed class IDs. 
However, their language understanding remains limited by relatively lightweight text encoders~\citep{devlin2019bert} and a narrow supervision interface dominated by region–text matching with category names or short noun phrases. 
As a result, they struggle to exploit heterogeneous vision-language supervision and remain limited in complex language instruction understanding and reasoning tasks. 

\textbf{MLLM-based Object Detection}~\citep{chen2023internvl, wang2025internvl3, guo2025seed1, bai2025qwen3, li2025visual, li2026unifying, jiang2026detect,wang2026locateanything} leverages pretrained large language models to unify heterogeneous vision-language tasks under a shared generative objective, enabling detection conditioned on complex semantics and language instructions. 
As an early precursor, Pix2Seq~\citep{chen2021pix2seq} quantizes object categories and box coordinates into discrete tokens and casts detection as auto-regressive sequence prediction. 
This formulation is also extended to MLLMs, allowing object detection, pointing, OCR, keypoint localization, and other vision-language tasks to be expressed through a shared generative interface. 
RexOmni~\citep{jiang2026detect} represents geometric locations using special coordinate tokens and adopts large-scale supervised fine-tuning followed by geometry-aware GRPO post-training to improve localization accuracy and mitigate coordinate drift and duplicate predictions. 
LocateAnything~\citep{wang2026locateanything} further introduces parallel box decoding, which treats each bounding box or point as an atomic geometric unit, preserving intra-box coherence while reducing the serial latency of token-by-token coordinate generation. 
Despite these advances, these works still preserve the ever-growing raw answer history during auto-regressive generation and do not explicitly address long-horizon dense grounding, where accumulated coordinate tokens can dominate subsequent predictions and induce self-reinforcing box repetition loop. 

\section{Method}
\label{sec:3}

\begin{figure}[t]
   \centering
   \includegraphics[width=\linewidth]{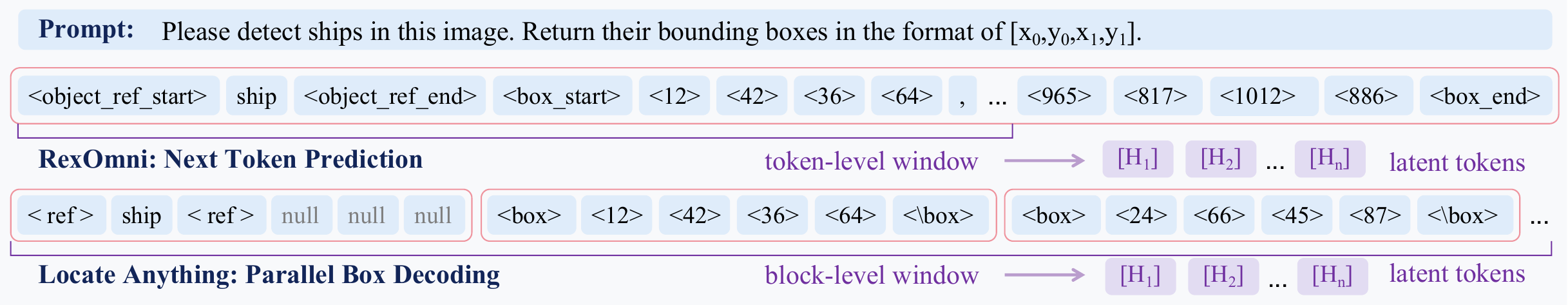}
   \caption{
      Illustration of the serialized response formats and compression granularities for RexOmni and LocateAnything.
      RexOmni represents the answer as a flat token sequence and generates it through next-token prediction, whereas LocateAnything organizes the answer into box-aligned blocks and predicts the coordinate tokens within each block in parallel.
      Accordingly, WinAR partitions RexOmni outputs into fixed-length token windows (e.g., every 300 tokens), while forming windows from a fixed number of complete blocks (e.g., every 50 blocks) for LocateAnything.
   }
   \label{fig3}
\end{figure}

We propose \textbf{Compressed Window Auto-Regression (WinAR)}, a simple and efficient paradigm for long-horizon vision-language grounding that replaces Full-History Autoregression with windowed generation conditioned on compact historical context. 
Section~\ref{sec:preliminaries} introduces the preliminaries of autoregressive vision-language grounding based on RexOmni~\citep{jiang2026detect} and LocateAnything~\citep{wang2026locateanything}, followed by a detailed description of WinAR in Section~\ref{sec:winar}. 

\subsection{Preliminaries}
\label{sec:preliminaries}

MLLM-based detectors are built upon pretrained multimodal large language models, such as Qwen2.5-VL-3B. 
Given an image $I$ and a natural-language prompt $P$, the model encodes $I$ into visual tokens $\mathbf{V}$ using a visual encoder~\citep{dosovitskiy2020image}, while the system instruction and prompt are tokenized and mapped into text embeddings. 
The resulting system, visual, and prompt tokens are concatenated to form the multimodal prefix
$\mathbf{X}=[\mathbf{S};\mathbf{V};\mathbf{P}]$. 
Under Full-History Auto-Regression, the target answer
$\mathbf{Y}=(y_1,\ldots,y_T)$
is generated through next-token prediction conditioned on the multimodal prefix $\mathbf{X}$ and the complete raw answer history:
\begin{equation}
    p_{\theta}(\mathbf{Y}\mid\mathbf{X})
    =
    \prod_{t=1}^{T}
    p_{\theta}
    \left(
        y_t \mid \mathbf{X},\mathbf{Y}_{<t}
    \right),
    \label{eq:autoregressive_generation}
\end{equation}
where $\mathbf{Y}_{<t}$ denotes all previously generated answer tokens. 
During supervised fine-tuning, the ground-truth answer is appended to the multimodal prefix $\mathbf{X}$ and optimized with cross-entropy loss under a causal attention mask. 
At inference time, the model first prefills the multimodal prefix $\mathbf{X}$ into the KV cache, and then autoregressively generates answer tokens while incrementally updating the cache until an end-of-sequence token or the maximum decoding length is reached. 

To adapt the general MLLM generation paradigm to object detection, RexOmni~\citep{jiang2026detect} and LocateAnything~\citep{wang2026locateanything} recast object detection as structured sequence prediction over category and coordinate tokens. 
Both methods encode spatial locations using quantized relative coordinates. 
Horizontal and vertical pixel coordinates are normalized by the image width and height, respectively, and uniformly quantized into $1{,}000$ discrete bins. 
The final $1{,}000$ entries of the language-model vocabulary, denoted by $\langle 0\rangle$ through $\langle 999\rangle$, are repurposed as coordinate tokens, allowing a bounding box $[x_0,y_0,x_1,y_1]$ to be represented by four coordinate tokens. 
As illustrated in Fig.~\ref{fig3}, RexOmni further uses the original structural tokens of Qwen2.5-VL-3B, such as \texttt{<|object\_ref\_start|>} and \texttt{<|box\_end|>}, to delimit each category phrase and its associated boxes, and serializes the complete answer into a flat sequence for standard next-token prediction. 
LocateAnything instead reorganizes the response into fixed-length atomic blocks of six tokens to support parallel box decoding. 
Different block types are used to represent category phrases and box outputs, while unused positions are padded with \texttt{<null>} tokens to maintain a uniform block length. 
Under this formulation, each box can be decoded in parallel within a single forward pass, reducing decoding latency while ensuring box-aligned and structural outputs. 

\subsection{Compressed Window Auto-Regression}
\label{sec:winar}

\begin{figure}[t]
   \centering
   \includegraphics[width=\linewidth]{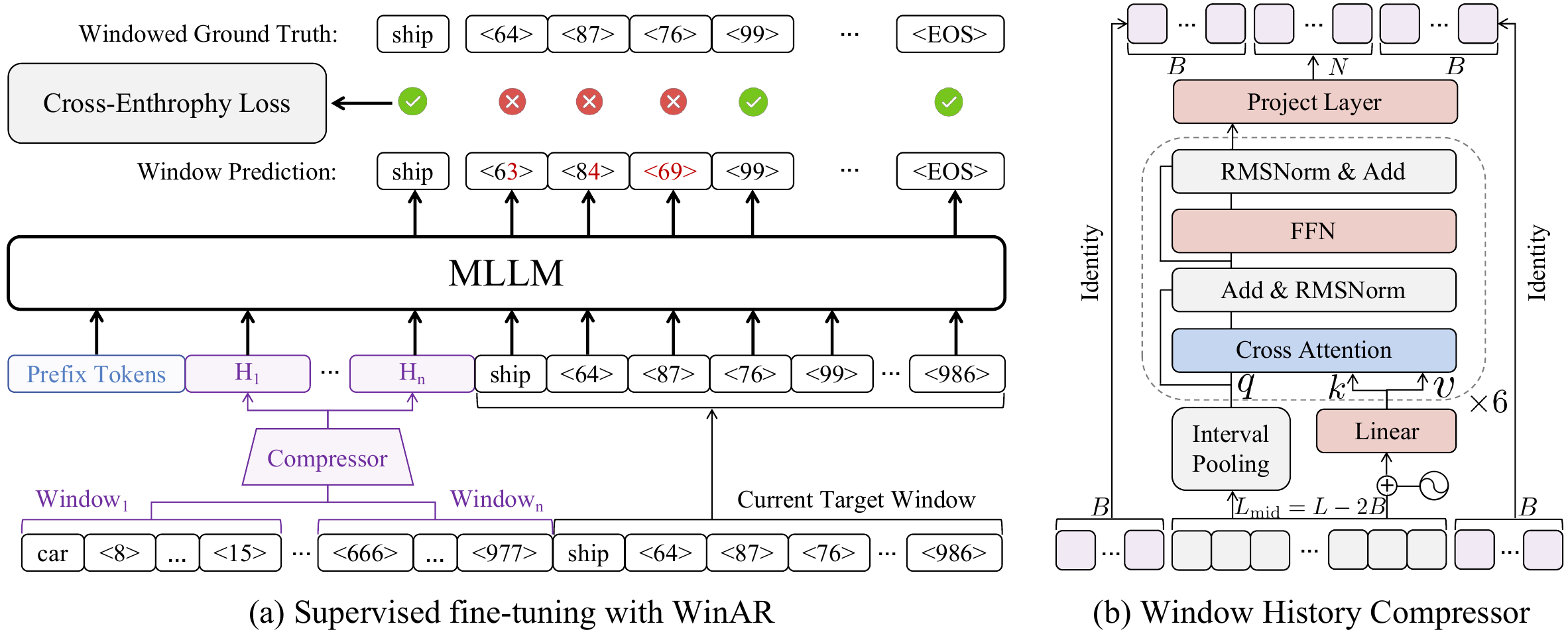}
   \caption{
      Overview of the proposed Compressed Window Auto-Regression.
      \textbf{(a)} A long answer is partitioned into consecutive windows and processed sequentially. 
      Preceding windows are compressed into memory and combined with the multimodal prefix to condition the current window, on which cross-entropy loss is applied. 
      \textbf{(b)} Structure of the Window History Compressor. 
   }
   \label{fig4}
\end{figure}

Motivated by the historical-answer attention takeover and box repetition loop analyzed in Section~\ref{sec:1}, WinAR replaces the ever-growing raw answer history of FHAR with windowed generation conditioned on compact historical context. 
As illustrated in Fig.~\ref{fig4}(a), WinAR follows a simple principle: \emph{remember compactly and generate locally}. 
Given a target answer $\mathbf{Y}=(y_1,\ldots,y_T)$ of length $T$ and a predefined window length $L$, we partition $\mathbf{Y}$ into
$K=\lceil T/L\rceil$ consecutive windows, where each window contains $L$ tokens except the last: 
\begin{equation}
    \mathbf{Y}
    =
    [\mathbf{Y}^{(1)};\mathbf{Y}^{(2)};\ldots;\mathbf{Y}^{(K)}],
\end{equation}
where $\mathbf{Y}^{(k)}$ denotes the $k$-th window. 
Before generating the $k$-th window, each completed window is independently transformed by the Window History Compressor into latent tokens: 
\begin{equation}
    \mathbf{H}^{(i)}
    =
    \mathcal{C}_{\phi}
    \left(
        \mathbf{Y}^{(i)}
    \right),
    \qquad i<k,
    \label{eq:window_compression}
\end{equation}
where $\mathcal{C}_{\phi}$ denotes the Window History Compressor, which is detailed in the following paragraph. 
The resulting generation process is then formulated as: 
\begin{equation}
\begin{aligned}
    p_{\theta}(\mathbf{Y}^{(k)}_{t}\mid\mathbf{X})
    =
    \prod_{t=1}^{T_k}
    p_{\theta}
    \left(
        y_t^{(k)}
        \mid
        \mathbf{X},
        \mathbf{H}_{<k},
        \mathbf{Y}^{(k)}_{<t}
    \right),
    \label{eq:windowed_generation}
\end{aligned}
\end{equation}
where only the local history
$\mathbf{Y}^{(k)}_{<t}$
of the current window remains explicitly visible.
The original multimodal prefix $\mathbf{X}$ is retained throughout generation, while the answer context grows only through compact latent tokens rather than raw coordinate token sequences. 

\textbf{Window History Compressor} transforms each completed answer window of length $L$ into a compact set of latent tokens. 
Since fixed-length windowing may split a category phrase or a geometric structure at the window boundary, we preserve $B$ tokens at both ends of the window to retain their exact local context. 
Thus the $k$-th window is decomposed as
$
    \mathbf{Y}^{(k)}
    =
    \left[
        \mathbf{y}^{(k)}_{\mathrm{front}};
        \mathbf{y}^{(k)}_{\mathrm{mid}};
        \mathbf{y}^{(k)}_{\mathrm{back}}
    \right],
$
where the front and back portions each contain $B$ boundary tokens and are passed directly to the output without compression, while the middle portion contains the remaining $L_{\mathrm{mid}}=L-2B$ tokens and is the only part processed by the compressor. 
As illustrated in Fig.~\ref{fig4}(b), given a predefined number of compressed latent tokens $N$, the middle tokens are processed through two complementary branches. 
In the first branch, the $(L-2B)$ middle tokens are uniformly partitioned into $N$ contiguous intervals, and each interval is average-pooled to initialize one query. 
This content-aware initialization provides each query with a localized summary of a distinct portion of the window, rather than relying on randomly initialized queries. 
To enable equal partitioning, the hyperparameters $L$, $B$, and $N$ are chosen such that $L-2B$ is divisible by $N$. 
In the second branch, learnable window-local positional embeddings are added to preserve relative order within the window, followed by a linear projection that produces the keys and values for cross-attention. 

The queries are then refined through stacked cross-attention and feed-forward layers. 
The resulting tokens are normalized and passed through an output projection consisting of two linear layers with a GELU activation in between. 
Finally, the compressed latent tokens are concatenated with the front tokens $\mathbf{y}^{(k)}_{\mathrm{front}}$ and back tokens $\mathbf{y}^{(k)}_{\mathrm{back}}$. 
The compressor thus reduces an $L$-token window to $2B+N$ latent tokens, preserving exact boundary context while compactly summarizing the middle portion. 

\textbf{Training and Inference.}
During training, the ground-truth answer is processed window by window. 
For the $k$-th training stage, all preceding windows are represented solely by
$\mathbf{H}_{<k}$, while cross-entropy supervision is applied only to the current target window. 
Raw windows can be discarded after compression, avoiding the need to retain full-sequence activations throughout training and thereby substantially reducing GPU memory consumption. 
As illustrated in Fig.~\ref{fig3}, windows are formed using a fixed number of answer tokens for RexOmni, whereas for LocateAnything, window boundaries are aligned with complete atomic blocks. 

At inference time, the model first generates
$\mathbf{Y}^{(1)}$
from the multimodal prefix $\mathbf{X}$. 
Once the current window is completed, it is compressed into
$\mathbf{H}^{(1)}$
and removed from the explicit answer context.
The model then generates the next window conditioned on the original prefix, the accumulated memory tokens, and the locally generated tokens within the current window.
This process repeats until an end-of-sequence token or the maximum decoding budget is reached. 

\section{Experiment}
\label{sec:exp}

\subsection{Training Details and Evaluation Setup}

\textbf{Training Details.}
For RexOmni, all variants are trained for 4 epochs with a per-device batch size of 1 and 16 gradient-accumulation steps. 
The base learning rate is $4\times10^{-5}$, with a vision-tower learning rate of $4\times10^{-6}$. \textbf{RexOmni-WinAR} additionally optimizes the Window History Compressor with a learning rate of $4\times10^{-4}$ and adopts token-level windowing with $L=300$, preserving $B=6$ tokens at each boundary and compressing the remaining tokens into $N=8$ latent tokens. For polygon detection, we use a larger window configuration with $L=400$, $B=8$, and $N=12$. 
To control for the increased exposure of long samples introduced by window-wise training, we further construct a \textbf{RexOmni-FHAR*} baseline, where a sample containing $T$ answer tokens is oversampled approximately $\lceil T/L\rceil$ times, matching both the effective sample exposure and the total number of optimizer updates of RexOmni-WinAR. 
We also conduct GRPO experiments following the reward design of RexOmni, training for one epoch on the subset of training samples containing more than 200 objects. 
For LocateAnything, we adopt the same optimization settings as RexOmni and construct corresponding \textbf{FHAR*} and \textbf{WinAR} variants. Since LocateAnything operates on box-aligned atomic blocks, LocateAnything-WinAR adopts block-level windowing with \(L=50\) blocks (300 tokens), preserves \(B=1\) block (6 tokens) at each boundary, and compresses the remaining blocks into \(N=2\) latent blocks (12 tokens). All FHAR-based experiments are conducted on eight NVIDIA H100 GPUs, whereas all WinAR experiments are conducted on eight NVIDIA A40 GPUs. More implementation details are provided in Section~\ref{sec:supp_impl}.

\textbf{Evaluation Setup.}
We conduct experiments on DOTA2~\citep{xia2018dota} and SODA-A~\citep{cheng2023towards}, jointly training on the union of their training sets and evaluating on the union of their validation sets, denoted as \textbf{DOTA2 \& SODA-A}. Following RexOmni~\citep{jiang2026detect}, we report category-averaged Recall, Precision, and F$_1$ score at an IoU threshold of 0.5. For polygon detection, we train separate models on the DOTA2 and SODA-A training sets and evaluate them on their corresponding validation sets. 
To further evaluate performance under extremely dense conditions, we select all evaluation images in DOTA2 \& SODA-A containing more than 200 objects to construct the \textbf{RSDense} subset and report image-averaged Recall, Precision, and F$1$ score at an IoU threshold of 0.5. 
To quantify prediction redundancy, we further report the duplicate rate $R{\mathrm{dup}}=N_{\mathrm{dup}}/N_{\mathrm{pred}}$ and the over-prediction rate $R_{\mathrm{over}}=(N_{\mathrm{pred}}-N_{\mathrm{gt}})/N_{\mathrm{gt}}$, where $N_{\mathrm{dup}}$, $N_{\mathrm{pred}}$, and $N_{\mathrm{gt}}$ denote the numbers of duplicate predictions, total predictions, and ground-truth objects, respectively. $R_{\mathrm{dup}}$ measures the fraction of redundant predictions caused by box repetition, while $R_{\mathrm{over}}$ measures the relative excess of the total prediction count over the ground-truth count. Both metrics are computed per image and then averaged over the evaluation set, denoted as Mean $R_{\mathrm{dup}}$ and Mean $R_{\mathrm{over}}$. More details on dataset settings and evaluation protocols are provided in Sections~\ref{sec:benchmark} and~\ref{sec:supp_evaluation_protocol}.

\subsection{Main Results}

\begin{table*}[t]
   \centering
   \caption{
   Comparison on DOTA2 \& SODA-A and the extremely dense RSDense subset. 
   $\mathrm{R}$, $\mathrm{P}$, and $\mathrm{F1}$ denote recall, precision, and F1 score at an IoU threshold of 0.5. 
   Mean $R_{\mathrm{dup}}$ and Mean $R_{\mathrm{over}}$ measure duplicate predictions and over-prediction, respectively. 
   The best and second-best results within each comparison group are highlighted in \textbf{bold} and \underline{underline}, respectively. 
   NTP denotes next token prediction, PBD denotes parallel box decoding, and Hybrid denotes a mixed decoding mode combining NTP and PBD. 
   See Locate Anything~\citep{wang2026locateanything} for details. 
   }
   \label{tab:main_results}

   \setlength{\tabcolsep}{4.2pt}
   \renewcommand{\arraystretch}{1.15}
   \small

   \resizebox{\textwidth}{!}{
   \begin{tabular}{
       l
       c
       ccc
       cc
       ccc
       cc
   }
   \toprule

   \multirow{3}{*}{Method}
   &
   \multirow{3}{*}{Mode}
   &
   \multicolumn{5}{c}{DOTA2 \& SODA-A}
   &
   \multicolumn{5}{c}{RSDense}
   \\

   \cmidrule(lr){3-7}
   \cmidrule(lr){8-12}

   &
   &
   \makecell{R@IoU$\uparrow$\\0.5}
   &
   \makecell{P@IoU$\uparrow$\\0.5}
   &
   \makecell{F1@IoU$\uparrow$\\0.5}
   &
   \makecell{Mean\\$R_{\mathrm{dup}}\downarrow$}
   &
   \makecell{Mean\\$R_{\mathrm{over}}\downarrow$}
   &
   \makecell{R@IoU$\uparrow$\\0.5}
   &
   \makecell{P@IoU$\uparrow$\\0.5}
   &
   \makecell{F1@IoU$\uparrow$\\0.5}
   &
   \makecell{Mean\\$R_{\mathrm{dup}}\downarrow$}
   &
   \makecell{Mean\\$R_{\mathrm{over}}\downarrow$}
   \\

   \midrule


   \multicolumn{12}{c}{\textbf{RexOmni}~\citep{jiang2026detect}}
   \\
   \midrule

   Baseline
   & NTP
   & 54.90
   & 62.56
   & 58.48
   & --
   & --
   & 17.32
   & 21.68
   & 12.69
   & --
   & --
   \\

   FHAR
   & NTP
   & 63.86
   & 61.66
   & 62.74
   & 4.58\%
   & 3.24
   & 23.39
   & 24.99
   & 16.57
   & 58.74\%
   & 5.85
   \\

   FHAR*
   & NTP
   & \underline{66.94}
   & \textbf{68.26}
   & \underline{67.59}
   & 2.43\%
   & \underline{1.39}
   & 36.82
   & \underline{37.92}
   & \underline{32.16}
   & 35.48\%
   & 3.14
   \\

   FHAR*-GRPO
   & NTP
   & 66.61
   & \underline{67.77}
   & 67.18
   & \underline{2.31\%}
   & 1.71
   & \underline{37.27}
   & 37.24
   & 31.98
   & \underline{35.05\%}
   & \underline{3.11}
   \\

   \rowcolor{blue!6}
   WinAR
   & NTP
   & \textbf{68.32}
   & 67.63
   & \textbf{67.97}
   & \textbf{1.26\%}
   & \textbf{0.51}
   & \textbf{45.23}
   & \textbf{38.90}
   & \textbf{39.36}
   & \textbf{13.00\%}
   & \textbf{0.87}
   \\

   \midrule


   \multicolumn{12}{c}{\textbf{Locate Anything}~\citep{wang2026locateanything}}
   \\
   \midrule


   Baseline
   & NTP
   & 27.68
   & 46.37
   & 34.66
   & --
   & --
   & 8.25
   & 32.41
   & 10.52
   & --
   & --
   \\

   FHAR*
   & NTP
   & \underline{68.96}
   & \textbf{71.52}
   & \textbf{70.34}
   & \underline{0.23\%}
   & \underline{0.19}
   & \underline{53.87}
   & \textbf{57.88}
   & \textbf{54.87}
   & \underline{5.90\%}
   & \underline{0.21}
   \\

   \rowcolor{blue!6}
   WinAR
   & NTP
   & \textbf{69.98}
   & \underline{68.08}
   & \underline{69.01}
   & \textbf{0.01\%}
   & \textbf{0.17}
   & \textbf{55.42}
   & \underline{56.22}
   & \underline{54.40}
   & \textbf{0.49\%}
   & \textbf{0.20}
   \\

   \cmidrule(lr){1-12}


   Baseline
   & PBD
   & 16.45
   & 19.25
   & 17.74
   & --
   & --
   & 1.01
   & 10.97
   & 1.54
   & --
   & --
   \\

   FHAR*
   & PBD
   & \underline{59.98}
   & \textbf{69.02}
   & \underline{64.18}
   & \underline{0.79\%}
   & \underline{0.39}
   & \underline{23.79}
   & \underline{32.47}
   & \underline{26.15}
   & \underline{14.89\%}
   & \underline{0.84}
   \\

   \rowcolor{blue!6}
   WinAR
   & PBD
   & \textbf{60.60}
   & \underline{68.74}
   & \textbf{64.41}
   & \textbf{0.35\%}
   & \textbf{0.15}
   & \textbf{23.98}
   & \textbf{34.44}
   & \textbf{27.54}
   & \textbf{2.13\%}
   & \textbf{0.02}
   \\

   \cmidrule(lr){1-12}


   Baseline
   & Hybrid
   & 26.48
   & 36.49
   & 30.69
   & --
   & --
   & 6.97
   & 25.39
   & 8.23
   & --
   & --
   \\

   FHAR*
   & Hybrid
   & \underline{67.95}
   & \underline{70.99}
   & \underline{69.38}
   & \underline{0.57\%}
   & \underline{0.24}
   & \underline{43.98}
   & \underline{55.27}
   & \underline{49.76}
   & \underline{10.60\%}
   & \underline{0.53}
   \\

   \rowcolor{blue!6}
   WinAR
   & Hybrid
   & \textbf{68.44}
   & \textbf{71.37}
   & \textbf{69.87}
   & \textbf{0.29\%}
   & \textbf{0.21}
   & \textbf{44.11}
   & \textbf{59.94}
   & \textbf{50.82}
   & \textbf{1.75\%}
   & \textbf{0.16}
   \\

   \bottomrule
   \end{tabular}
   }
\end{table*}

\begin{table}[t]
    \centering
    \caption{
        Comparison of inference speed on RSDense and performance on two extremely dense categories.\textbf{Boxes/s} denotes the average number of predicted boxes per image per second, reflecting the decoding throughput. \textbf{Valid Boxes/s} measures the throughput after removing redundant boxes. 
        Inference speed is measured on eight NVIDIA H100 GPUs. 
        }
    \label{tab:dense_categories_efficiency}
 
    \setlength{\tabcolsep}{4.5pt}
    \renewcommand{\arraystretch}{1.15}
    \small
 
    \resizebox{\columnwidth}{!}{
    \begin{tabular}{
        l
        ccc
        ccc
        ccc
    }
    \toprule

    \multirow{3}{*}{RexOmni}
    &
    \multirow{3}{*}{box  / s$\uparrow$}
    &
    \multirow{3}{*}{valid  box / s$\uparrow$}
    &
    \multirow{3}{*}{\makecell{Training \\ Peak GPU\\Memory (GB)$\downarrow$}}
    &
    \multicolumn{3}{c}{Ship}
    &
    \multicolumn{3}{c}{Small Vehicle}
    \\

    \cmidrule(lr){5-7}
    \cmidrule(lr){8-10}

    &
    &
    &
    &
    \makecell{R@IoU$\uparrow$\\0.5}
    &
    \makecell{P@IoU$\uparrow$\\0.5}
    &
    \makecell{F1@IoU$\uparrow$\\0.5}
    &
    \makecell{R@IoU$\uparrow$\\0.5}
    &
    \makecell{P@IoU$\uparrow$\\0.5}
    &
    \makecell{F1@IoU$\uparrow$\\0.5}
    \\
 
    \midrule
 
    \makecell[l]{FHAR*}
    & 6.05
    & 3.57
    & 57.62
    & 57.13
    & 47.23
    & 51.71
    & 22.65
    & 17.80
    & 19.94
    \\
 
    \rowcolor{blue!6}
    WinAR
    & \textbf{6.68}
    & \textbf{5.40}
    & \textbf{42.97}
    & \textbf{65.35}
    & \textbf{58.50}
    & \textbf{61.74}
    & \textbf{30.06}
    & \textbf{24.45}
    & \textbf{26.96}
    \\
 
    \bottomrule
    \end{tabular}
    }
 \end{table}

\begin{table*}[t]
   \centering
   \caption{
   Polygon detection results on DOTA2 and SODA-A. The Baseline is RexOmni model. 
   }
   \label{tab:polygon_results}

   \setlength{\tabcolsep}{4.2pt}
   \renewcommand{\arraystretch}{1.15}
   \small

   \resizebox{\textwidth}{!}{
   \begin{tabular}{
       l
       ccc
       cc
       ccc
       cc
   }
   \toprule

   \multirow{3}{*}{Method}
   &
   \multicolumn{5}{c}{DOTA2}
   &
   \multicolumn{5}{c}{SODA-A}
   \\

   \cmidrule(lr){2-6}
   \cmidrule(lr){7-11}

   &
   \makecell{R@IoU$\uparrow$\\0.5}
   &
   \makecell{P@IoU$\uparrow$\\0.5}
   &
   \makecell{F1@IoU$\uparrow$\\0.5}
   &
   \makecell{Mean\\$R_{\mathrm{dup}}\downarrow$}
   &
   \makecell{Mean\\$R_{\mathrm{over}}\downarrow$}
   &
   \makecell{R@IoU$\uparrow$\\0.5}
   &
   \makecell{P@IoU$\uparrow$\\0.5}
   &
   \makecell{F1@IoU$\uparrow$\\0.5}
   &
   \makecell{Mean\\$R_{\mathrm{dup}}\downarrow$}
   &
   \makecell{Mean\\$R_{\mathrm{over}}\downarrow$}
   \\

   \midrule

   Baseline
   & 2.49
   & 3.06
   & 2.44
   & --
   & --
   & 1.95
   & 2.27
   & 2.03
   & --
   & --
   \\

   FHAR
   & 46.79
   & 41.82
   & 39.67
   & 8.93\%
   & 8.03
   & 47.48
   & 27.19
   & 32.78
   & 7.16\%
   & 6.32
   \\

   FHAR*
   & \underline{51.20}
   & \underline{47.98}
   & \underline{47.25}
   & \underline{6.34\%}
   & \underline{4.99}
   & \underline{54.45}
   & \underline{34.71}
   & \underline{38.64}
   & \underline{3.72\%}
   & \underline{3.27}
   \\

   \rowcolor{blue!6}
   WinAR
   & \textbf{52.04}
   & \textbf{49.30}
   & \textbf{48.93}
   & \textbf{2.46\%}
   & \textbf{0.89}
   & \textbf{55.35}
   & \textbf{39.38}
   & \textbf{44.00}
   & \textbf{1.93\%}
   & \textbf{0.75}
   \\

   \bottomrule
   \end{tabular}
   }
\end{table*}

\textbf{Dense Grounding and Box Repetition Suppression.}
Table~\ref{tab:main_results} reports the main results, where the advantage of WinAR becomes increasingly pronounced as scene density increases. 
On RexOmni, WinAR improves Recall from 66.94 to 68.32 over the FHAR* baseline on DOTA2 \& SODA-A, while reducing Mean $R_{\mathrm{dup}}$ from 2.43\% to 1.26\% and Mean $R_{\mathrm{over}}$ from 1.39 to 0.51.
The improvement becomes substantially more pronounced on RSDense: WinAR increases Recall from 36.82 to 45.23 and F1 from 32.16 to 39.36, while reducing Mean $R_{\mathrm{dup}}$ from 35.48\% to 13.00\% and Mean $R_{\mathrm{over}}$ from 3.14 to 0.87. 
The same trend is particularly evident for densely populated categories.
As shown in Table~\ref{tab:dense_categories_efficiency}, WinAR improves the Recall of Ship from 57.13 to 65.35 and that of Small Vehicle from 22.65 to 30.06 on RSDense, with their F1 scores increasing from 51.71 to 61.74 and from 19.94 to 26.96, respectively. 
These improvements are also reflected in the qualitative comparisons in Fig.~\ref{fig1}(middle), where WinAR suppresses repeated boxes and recovers substantially more distinct objects. 
As shown in Fig.~\ref{fig2}(b), compressing the historical answer mitigates the progressive concentration of attention on the answer history and maintains more stable conditioning throughout decoding.
Consequently, WinAR suppresses the box repetition loop and preserves more of the finite decoding budget for discovering new objects, thereby sustaining object discovery over long decoding horizons. 
More experimental results across additional datasets are provided in Section~\ref{sec:more_experiments}. 

\textbf{Generalization across Decoding Paradigms.}
WinAR also generalizes to the parallel box decoding framework of LocateAnything, as shown in Table~\ref{tab:main_results}.
Across NTP, PBD, and Hybrid decoding, WinAR consistently suppresses redundant predictions while maintaining comparable or improved grounding accuracy. 
This effect is particularly evident on the RSDense benchmark.
Under NTP, WinAR reduces Mean $R_{\mathrm{dup}}$ from 5.90\% to 0.49\% on RSDense, while achieving comparable F1 (54.87 vs.\ 54.40) and improving Recall from 53.87 to 55.42. 
Under PBD, WinAR improves F1 from 26.15 to 27.54, while reducing Mean $R_{\mathrm{dup}}$ from 14.89\% to 2.13\% on RSDense. 
This demonstrates that WinAR is not tied to token-by-token box generation and can also effectively control historical-answer accumulation when geometric outputs are decoded in parallel. 

\textbf{Generalization to Other Structured Geometric Outputs.}
Table~\ref{tab:polygon_results} evaluates whether WinAR generalizes beyond bounding boxes to polygon detection, where each object is represented by a longer sequence of interdependent coordinates (eight coordinate tokens per polygon). 
Compared with the exposure-matched FHAR* baseline, WinAR improves F1 from 47.25 to 48.93 on DOTA2, while reducing Mean $R_{\mathrm{dup}}$ from 6.34\% to 2.46\% and Mean $R_{\mathrm{over}}$ from 4.99 to 0.89. 
The improvement is more pronounced on SODA-A, which contains generally denser scenes: Recall, Precision, and F1 increase from 54.45, 34.71, and 38.64 to 55.35, 39.38, and 44.00, respectively, while Mean $R_{\mathrm{dup}}$ decreases from 3.72\% to 1.93\% and Mean $R_{\mathrm{over}}$ from 3.27 to 0.75. 
These results demonstrate that WinAR is not specific to fixed-length bounding-box sequences, but also effectively controls historical-answer accumulation when generating more complex structured geometric outputs. 

\textbf{Training and Inference Efficiency.}
WinAR substantially reduces the memory requirement during training.
In our training setup, the peak GPU memory of RexOmni-FHAR* reaches 57.62~GB, whereas RexOmni-WinAR requires only 42.97~GB. 
This reduction stems from the window-wise training formulation, where completed answer windows are compressed and their raw token-level representations and associated activations can be discarded, substantially reducing the per-GPU memory budget and enabling all WinAR experiments to be conducted on 48~GB NVIDIA A40 GPUs. 
Table~\ref{tab:dense_categories_efficiency} also evaluates the inference efficiency of WinAR on RSDense. 
WinAR does not consistently improve raw throughput due to the additional overhead of history compression and KV-cache prefilling at window boundaries. However, by substantially reducing redundant predictions, it uses the decoding budget more effectively and increases Valid Boxes/s from 3.57 to 5.40 on RexOmni. 
Overall, WinAR improves both training scalability and effective inference throughput by replacing ever-growing raw answer histories with compact window-level representations.

\subsection{Ablation Study}

\begin{table*}[t]
   \centering
   \caption{
   Ablation study of WinAR with different window length $L$, boundary size $B$, and number of compressed memory tokens $N$. 
   }
   \label{tab:compression_ablation}

   \setlength{\tabcolsep}{4.2pt}
   \renewcommand{\arraystretch}{1.15}
   \small

   \resizebox{\textwidth}{!}{
   \begin{tabular}{
       ccc
       ccc
       cc
       ccc
       cc
   }
   \toprule

   \multicolumn{3}{c}{WinAR}
   &
   \multicolumn{5}{c}{DOTA2 \& SODA-A}
   &
   \multicolumn{5}{c}{RSDense}
   \\

   \cmidrule(lr){1-3}
   \cmidrule(lr){4-8}
   \cmidrule(lr){9-13}

   $L$
   &
   $B$
   &
   $N$
   &
   \makecell{R@IoU$\uparrow$\\0.5}
   &
   \makecell{P@IoU$\uparrow$\\0.5}
   &
   \makecell{F1@IoU$\uparrow$\\0.5}
   &
   \makecell{Mean\\$R_{\mathrm{dup}}\downarrow$}
   &
   \makecell{Mean\\$R_{\mathrm{over}}\downarrow$}
   &
   \makecell{R@IoU$\uparrow$\\0.5}
   &
   \makecell{P@IoU$\uparrow$\\0.5}
   &
   \makecell{F1@IoU$\uparrow$\\0.5}
   &
   \makecell{Mean\\$R_{\mathrm{dup}}\downarrow$}
   &
   \makecell{Mean\\$R_{\mathrm{over}}\downarrow$}
   \\

   \midrule

   300
   & 6
   & 8
   & \textbf{68.67}
   & \textbf{68.41}
   & \textbf{68.49}
   & 1.21\%
   & 0.48
   & \textbf{48.59}
   & \textbf{42.82}
   & \textbf{44.78}
   & 5.38\%
   & 0.66
   \\

   200
   & 4
   & 6
   & 67.39
   & \underline{67.61}
   & \underline{67.22}
   & 0.65\%
   & 0.42
   & \underline{47.93}
   & \underline{42.60}
   & \underline{44.56}
   & 4.21\%
   & 0.57
   \\

   200
   & 0
   & 20
   & 63.40
   & 67.16
   & 64.82
   & \underline{0.28\%}
   & \textbf{0.18}
   & 27.96
   & 39.59
   & 32.47
   & \underline{2.59\%}
   & \textbf{0.04}
   \\

   100
   & 2
   & 4
   & \underline{68.40}
   & 66.15
   & 67.04
   & \textbf{0.12\%}
   & \underline{0.27}
   & 47.04
   & 41.41
   & 43.71
   & \textbf{0.89\%}
   & \underline{0.32}
   \\

   \bottomrule
   \end{tabular}
   }
\end{table*}

\begin{table*}[t]
   \centering
   \caption{
   Ablation study on the number of layers in the Window History Compressor. 
   }
   \label{tab:layer_ablation}

   \setlength{\tabcolsep}{4.2pt}
   \renewcommand{\arraystretch}{1.15}
   \small

   \resizebox{\textwidth}{!}{
   \begin{tabular}{
       c
       ccc
       cc
       ccc
       cc
   }
   \toprule

   \multirow{2}{*}{\makecell{Compressor\\Layers}}
   &
   \multicolumn{5}{c}{DOTA2 \& SODA-A}
   &
   \multicolumn{5}{c}{RSDense}
   \\

   \cmidrule(lr){2-6}
   \cmidrule(lr){7-11}

   &
   \makecell{R@IoU$\uparrow$\\0.5}
   &
   \makecell{P@IoU$\uparrow$\\0.5}
   &
   \makecell{F1@IoU$\uparrow$\\0.5}
   &
   \makecell{Mean\\$R_{\mathrm{dup}}\downarrow$}
   &
   \makecell{Mean\\$R_{\mathrm{over}}\downarrow$}
   &
   \makecell{R@IoU$\uparrow$\\0.5}
   &
   \makecell{P@IoU$\uparrow$\\0.5}
   &
   \makecell{F1@IoU$\uparrow$\\0.5}
   &
   \makecell{Mean\\$R_{\mathrm{dup}}\downarrow$}
   &
   \makecell{Mean\\$R_{\mathrm{over}}\downarrow$}
   \\

   \midrule

   8
   & 66.17
   & 63.14
   & 64.62
   & 1.31\%
   & 0.66
   & \textbf{45.75}
   & \underline{38.23}
   & \underline{39.32}
   & 16.87\%
   & 1.28
   \\

   6
   & \textbf{68.32}
   & \textbf{67.63}
   & \textbf{67.97}
   & \underline{1.26\%}
   & \underline{0.51}
   & \underline{45.23}
   & \textbf{38.90}
   & \textbf{39.36}
   & \textbf{13.00\%}
   & \underline{0.87}
   \\

   4
   & \underline{66.57}
   & \underline{64.45}
   & \underline{65.49}
   & \textbf{1.23\%}
   & \textbf{0.44}
   & 43.60
   & 38.20
   & 37.73
   & \underline{15.64\%}
   & \textbf{0.78}
   \\

   \bottomrule
   \end{tabular}
   }
\end{table*}

\textbf{Compression Configuration.} 
Table~\ref{tab:compression_ablation} studies different compression configurations by varying the window length $L$, boundary size $B$, and number of compressed tokens $N$. 
Since a smaller window length naturally induces more window-wise training instances, the $L=100$ configuration receives a larger oversampling weight. 
To ensure a fair comparison, we apply the same oversampling weight to the $L=200$ and $L=300$ configurations, thereby matching their effective sample exposure. 
We make two main observations. 
First, more frequent compression consistently suppresses redundant generation, but introduces a trade-off with grounding accuracy. 
Reducing $L$ from 300 to 100 decreases Mean $R_{\mathrm{dup}}$ from 5.38\% to 0.89\% and Mean $R_{\mathrm{over}}$ from 0.66 to 0.32 on RSDense, indicating that more frequent compression more effectively limits the accumulation of raw answer history. 
However, the corresponding F1 decreases from 44.78 to 43.71, with both Recall and Precision slightly reduced. 
This suggests that overly short windows, while effective at suppressing repetition, provide less exact local context for subsequent generation. 
Second, preserving exact boundary tokens is important for maintaining grounding performance. 
With the same window length $L=200$, removing boundary preservation ($B=0$) and increasing the number of compressed tokens to $N=20$ reduces F1 from 67.22 to 64.82 on DOTA2 \& SODA-A and, more substantially, from 44.56 to 32.47 on RSDense compared with $(B,N)=(4,6)$. 
This degradation occurs despite the considerably larger compressed-token budget, indicating that simply allocating more latent tokens cannot compensate for the loss of exact local information around window boundaries. 
Overall, $(L,B,N)=(300,6,8)$ achieves the best grounding accuracy while still substantially suppressing redundant generation, and is therefore adopted as our default configuration. 

\textbf{Compressor Depth.}
Table~\ref{tab:layer_ablation} studies the effect of the depth in the Window History Compressor. 
A shallow 4-layer compressor is less effective at preserving informative historical context in extremely dense scenes. 
Increasing the depth to 6 layers improves the overall compression quality, yielding the best F1 on both DOTA2 \& SODA-A (67.97) and RSDense (39.36), while also achieving the lowest Mean $R_{\mathrm{dup}}$ of 13.00\% on RSDense. 
Further increasing the depth to 8 layers slightly improves Recall on RSDense from 45.23 to 45.75, but decreases Precision from 38.90 to 38.23 and increases Mean $R_{\mathrm{dup}}$ from 13.00\% to 16.87\%, resulting in a slightly lower F1 of 39.32. 
These results suggest that insufficient depth limits the compressor's capacity to retain useful historical information, whereas additional layers beyond six provide little overall benefit and may introduce unnecessary transformations. 
We therefore adopt a 6-layer Window History Compressor by default.

\section{Conclusions}

In this work, we identify a critical long-horizon failure mode of MLLM-based detectors in dense vision-language grounding, where generation progressively collapses into a self-reinforcing box repetition loop. 
To address this issue, we propose \textbf{Compressed Window Auto-Regression (WinAR)}, which preserves exact autoregression within the current window while representing completed windows with compact latent tokens, thereby limiting repeated exposure to raw historical coordinates without discarding cross-window dependencies. 
Experiments show that WinAR increasingly benefits denser scenes. WinAR further generalizes across decoding paradigms and structured geometric outputs, while improving effective detection efficiency and reducing training memory for long sequences. 
These results establish compressed window autoregression as a practical paradigm for reliable long-horizon vision-language grounding.



\bibliography{iclr2027_conference}
\bibliographystyle{iclr2027_conference}

\newpage
\appendix
\section{Appendix}

\subsection*{Appendix Contents}
\begin{itemize}
    \item \textbf{\hyperref[sec:benchmark]{Benchmark Details}} summarizes the construction, preprocessing, splits, and statistics of DOTA2 \& SODA-A, RSDense, DOTA2-Poly, SODA-A-Poly, SKU110K, and COCO.
    \item \textbf{\hyperref[sec:supp_impl]{Additional Implementation Details}} provides the training configurations for RexOmni and LocateAnything.
    \item \textbf{\hyperref[sec:supp_evaluation_protocol]{Evaluation Protocol}} describes the capped evaluation protocol and the complementary all-prediction evaluation for analyzing object discovery, prediction redundancy, and generation termination.
    \item \textbf{\hyperref[sec:more_experiments]{More Experiments}} reports all-prediction comparisons across COCO, DOTA2 \& SODA-A, SKU110K, and RSDense.
\end{itemize}

\subsection{Benchmark Details}
\label{sec:benchmark}

In this section, we summarize the datasets and benchmark settings used throughout our experiments. We evaluate WinAR from three complementary perspectives. First, we use DOTA2 \& SODA-A~\citep{xia2018dota, cheng2023towards}, RSDense, and SKU110K~\citep{goldman2019precise} to assess performance in dense and extremely dense scenes across both remote-sensing and general visual domains. Second, we construct DOTA2-Poly~\citep{xia2018dota} and SODA-A-Poly~\citep{cheng2023towards} to evaluate long-horizon structured generation with quadrilateral outputs, where each object requires twice as many coordinate tokens as a standard bounding box. Finally, we use COCO~\citep{lin2014microsoft} to verify that the improvements of WinAR in dense scenes do not compromise performance in ordinary visual scenarios.

\begin{itemize}
    \item \textbf{DOTA2 \& SODA-A: }DOTA2~\citep{xia2018dota} and SODA-A~\citep{cheng2023towards} are both remote-sensing object detection datasets with closely aligned scene domains and a substantial number of densely populated images. We therefore combine the two datasets to construct a unified training and evaluation benchmark for studying dense vision-language grounding across different scene densities.
    For DOTA-v2.0, we follow the official single-scale preprocessing and evaluation protocol. Specifically, the original high-resolution images are cropped into $1024\times1024$ patches with an overlap of 200 pixels between adjacent patches. The resulting training split, containing 13,447 images, is included in the unified training set, while the validation split, containing 4,053 images, is used as part of the unified test set.
    For SODA-A, we follow the official preprocessing setting and crop the original images into $800\times800$ patches with a stride of 650 pixels. We merge its training and validation splits, yielding 32,181 images for training, while the test split, containing 14,654 images, is included in the unified test set.
    After preprocessing, the DOTA2 and SODA-A training data are jointly used for model training, while the DOTA2 validation split and the SODA-A test split are combined to form the DOTA2 \& SODA-A evaluation benchmark used throughout this work.
    \item \textbf{RSDense: }To specifically evaluate WinAR in highly dense visual scenes, we construct RSDense from the DOTA2~\citep{xia2018dota} \& SODA-A~\citep{cheng2023towards} evaluation set by selecting images containing more than 200 objects. RSDense consists of 712 images, with an average of 412 objects per image, providing a challenging benchmark for long-horizon dense grounding. Among all categories, \textit{Ship} and \textit{Small Vehicle} exhibit the highest object densities and therefore serve as two representative dense categories in our analysis. 
    \item \textbf{DOTA2-Poly and SODA-A-Poly: }We evaluate WinAR on the oriented-box annotations of DOTA2~\citep{xia2018dota} and SODA-A~\citep{cheng2023towards} to assess its generalization to quadrilateral geometric outputs. Unlike a standard horizontal box, which is represented by four coordinate tokens, an oriented box is described by four vertices and therefore requires eight coordinate tokens. For the same number of objects, this doubles the output length and substantially extends the decoding horizon, imposing greater demands on long-range autoregressive modeling and historical-context management. This setting therefore provides a more challenging test of WinAR under longer and more complex structured generation.
    In these experiments, we train and evaluate on the two datasets separately rather than combining them. For DOTA2-Poly, we use the oriented-box annotations of DOTA2, train on its training split, and evaluate on the corresponding validation split. For SODA-A-Poly, we use the oriented-box annotations of SODA-A, merge its training and validation splits for training, and evaluate on the test split. All other preprocessing and training settings are kept consistent with their horizontal-box counterparts, with only the output representation changed from 4-token box coordinates to 8-token quadrilateral coordinates.
    \item \textbf{SKU110K: }The SKU110K dataset~\citep{goldman2019precise} is a large-scale object detection dataset focused on densely packed retail products on supermarket shelves, representing a challenging dense-scene setting in general visual domains. We use its training split, containing 8,769 images, for model training and evaluate on the test split, which contains 2,920 images. This benchmark is used to assess whether WinAR generalizes beyond remote-sensing imagery to dense object detection in every scenes. 
    \item \textbf{COCO: }To verify that the gains of WinAR in dense scenes do not come at the expense of performance in ordinary visual scenarios, we further evaluate it on the widely used COCO~\citep{lin2014microsoft} benchmark. Following standard practice~\citep{jiang2026detect}, we train on the COCO training set and evaluate on the validation set. This setting allows us to assess whether WinAR preserves competitive grounding performance under typical scene densities while improving robustness in highly dense scenes.
\end{itemize}

\begin{table}[h]
    \centering
    \caption{
        Detailed statistics of the training datasets used in our experiments.
    }
    \label{tab:training_dataset_statistics}

    \setlength{\tabcolsep}{10pt}
    \renewcommand{\arraystretch}{1.15}
    \small

    \resizebox{\columnwidth}{!}{
    \begin{tabular}{lccccc}
        \toprule
        \textbf{Dataset}
        & \textbf{\#Images}
        & \textbf{\#Annotations}
        & \textbf{Min / Image}
        & \textbf{Max / Image}
        & \textbf{Avg. / Image}
        \\
        \midrule

        DOTA2 \& SODA-A
        & 45,628
        & 1,493,991
        & 1
        & 2,831
        & 32.74
        \\

        DOTA2-Poly
        & 13,447
        & 473,548
        & 1
        & 2,831
        & 35.22
        \\

        SODA-A-Poly
        & 32,181
        & 1,020,443
        & 1
        & 2,653
        & 31.71
        \\

        SKU110K
        & 8,769
        & 1,293,724
        & 1
        & 718
        & 147.53
        \\

        COCO
        & 117,266
        & 860,001
        & 1
        & 93
        & 7.33
        \\

        \bottomrule
    \end{tabular}
    }
\end{table}

\begin{table}[h]
    \centering
    \caption{
        Detailed statistics of the test datasets used in our experiments.
    }
    \label{tab:evaluation_dataset_statistics}

    \setlength{\tabcolsep}{6pt}
    \renewcommand{\arraystretch}{1.15}
    \small

    \resizebox{\columnwidth}{!}{
    \begin{tabular}{lccccc}
        \toprule
        \textbf{Dataset}
        & \textbf{\#Images}
        & \textbf{\#Annotations}
        & \textbf{Min Objects/Image}
        & \textbf{Max Objects/Image}
        & \textbf{Avg. Objects/Image}
        \\
        \midrule

        DOTA2 \& SODA-A
        & 18,707
        & 656,706
        & 1
        & 2,305
        & 35.14
        \\

        RSDense
        & 712
        & 293,371
        & 201
        & 2,305
        & 412.04
        \\

        SKU110K
        & 2,920
        & 429,411
        & 20
        & 533
        & 147.06
        \\

        DOTA2-Poly
        & 4,053
        & 140,103
        & 1
        & 2,307
        & 34.57
        \\

        SODA-A-Poly
        & 14,654
        & 517,099
        & 1
        & 1,692
        & 35.29
        \\

        COCO
        & 4,952
        & 36,781
        & 1
        & 63
        & 7.43
        \\

        \bottomrule
    \end{tabular}
    }
\end{table}

\subsection{Additional Implementation Details}
\label{sec:supp_impl}

Table~\ref{tab:training_details} summarizes the training configurations for RexOmni~\citep{jiang2026detect} and LocateAnything~\citep{wang2026locateanything}. 
A key consideration in our comparison is that window-wise training in WinAR naturally increases both the number of optimizer updates and the effective exposure of long-answer samples.
To disentangle the benefit of WinAR from that of additional training exposure, we carefully construct three training settings:

\begin{itemize}
    \item \textbf{RexOmni-FHAR.}We first reproduce the original Full-History Auto-Regression (FHAR) training protocol of RexOmni. 
    The model is trained for 4 epochs with a per-device batch size of 1 and 16 gradient-accumulation steps, resulting in 1,424 optimizer updates. 
    We use AdamW with a base learning rate of $4\times10^{-5}$, an MM-projector learning rate of $4\times10^{-5}$, and a vision-tower learning rate of $4\times10^{-6}$. 
    The learning rate follows a cosine schedule with a warmup ratio of 0.05. 
    We set the weight decay to 0.05, the maximum gradient norm to 1.0, and the maximum sequence length to 20,000 tokens.
    Each training sample is exposed once per epoch, regardless of its answer length. 
    \item \textbf{RexOmni-WinAR.}RexOmni-WinAR follows the same optimization settings as RexOmni-FHAR, while additionally optimizing the Window History Compressor with a learning rate of $4\times10^{-4}$. 
    WinAR adopts a window size of $L=300$ and decomposes each long answer into multiple consecutive windows that are optimized separately. 
    Consequently, long-answer samples receive proportionally more training exposure than under conventional FHAR. 
    Specifically, a sample containing $T$ answer tokens contributes approximately $\lceil T/L\rceil$ window-level training instances; for example, an answer of 3,000 tokens is divided into 10 windows when $L=300$. 
    This window-wise formulation therefore implicitly oversamples long samples and increases the number of training iterations within each epoch. 
    With our training set, RexOmni-WinAR performs 1,864 optimizer updates over 4 epochs. 
    \item \textbf{RexOmni-FHAR$^*$.}
    To isolate the effect of compressed window autoregression from the additional sample exposure introduced by window-wise training, we construct an exposure-matched FHAR$^*$ baseline. 
    FHAR$^*$ retains the original full-history autoregressive training objective, but modifies data sampling such that a sample with $T$ answer tokens is repeated approximately $\lceil T/L\rceil$ times, matching the effective exposure assigned to that sample by RexOmni-WinAR. 
    This results in the same 1,864 optimizer updates as WinAR over 4 epochs, while preserving the FHAR decoding context during training. 
    All remaining optimization hyperparameters are kept identical to RexOmni-FHAR. 
    \item \textbf{RexOmni-FHAR$^*$-GRPO.}
    Post-training with GRPO is an important component of RexOmni for improving detection quality. 
    In our experiments, GRPO is applied to the RexOmni baseline for comparison with the original RexOmni training recipe.
    We initialize the policy from the SFT checkpoint of RexOmni-FHAR$^*$ and perform GRPO post-training for one epoch on the subset of the joint DOTA2 and SODA-A training set containing more than 200 objects per image. 
    Training is distributed over 8 NVIDIA H100 GPUs.
    For each rollout iteration, we sample a batch of 8 prompts  and generate 4 rollouts for each prompt. 
    The four candidate trajectories associated with the same prompt form a response group for group-relative reward comparison in GRPO, resulting in 32 sampled responses per rollout batch.
    The actor is optimized with a global batch size of 8.
    \item \textbf{LocateAnything-FHAR$^*$.}
    We follow the same training strategy as RexOmni-FHAR$^*$.
    To account for the increased sample exposure introduced by window-wise training in LocateAnything-WinAR, we oversample the corresponding training samples to match its effective sample exposure and number of training iterations.
    All other optimization hyperparameters are kept consistent with those used for RexOmni.
    \item \textbf{LocateAnything-WinAR.}
    LocateAnything-WinAR follows the same training strategy as RexOmni-WinAR, except that windows are constructed over box-aligned atomic blocks rather than individual tokens.
    Specifically, each long answer is partitioned into consecutive block-level windows, which generates additional training instances and consequently increases both the number of iterations per epoch and the exposure frequency of samples spanning multiple windows.
    The remaining optimization settings, including the learning rate of the Window History Compressor, are kept consistent with RexOmni-WinAR.
\end{itemize}

Therefore, FHAR$^*$ matches WinAR in both the total number of optimizer updates and the effective exposure of long-answer samples, while differing only in how the historical answer is represented during autoregressive training. 
This controlled comparison allows us to distinguish the effect of WinAR from gains that may arise simply from increased training exposure. 

\begin{table}[h]
    \centering
    \caption{
        Training hyperparameters for RexOmni and LocateAnything.
        RO and LA denote RexOmni and LocateAnything, respectively.
    }
    \label{tab:training_details}

    \setlength{\tabcolsep}{6pt}
    \renewcommand{\arraystretch}{1.15}
    \small

    \resizebox{\columnwidth}{!}{
    \begin{tabular}{lccccc}
        \toprule
        \textbf{Hyperparameter}
        & \textbf{RO-FHAR}
        & \textbf{RO-FHAR*}
        & \textbf{RO-WinAR}
        & \textbf{LA-FHAR*}
        & \textbf{LA-WinAR}
        \\
        \midrule

        Training Epochs
        & 4
        & 4
        & 4
        & 4
        & 4
        \\

        Oversample
        & False
        & True
        & True
        & True
        & True
        \\

        Optimize Steps
        & 1,424
        & 1,864
        & 1,864
        & 2,004
        & 2,004
        \\

        Per-device Batch Size
        & 1
        & 1
        & 1
        & 1
        & 1
        \\

        Gradient Accumulation Steps
        & 16
        & 16
        & 16
        & 16
        & 16
        \\

        Base Learning Rate
        & $4\times10^{-5}$
        & $4\times10^{-5}$
        & $4\times10^{-5}$
        & $4\times10^{-5}$
        & $4\times10^{-5}$
        \\

        MM Projector Learning Rate
        & $4\times10^{-5}$
        & $4\times10^{-5}$
        & $4\times10^{-5}$
        & $4\times10^{-5}$
        & $4\times10^{-5}$
        \\

        Vision Tower Learning Rate
        & $4\times10^{-6}$
        & $4\times10^{-6}$
        & $4\times10^{-6}$
        & $4\times10^{-6}$
        & $4\times10^{-6}$
        \\

        Window History Compressor LR
        & --
        & --
        & $4\times10^{-4}$
        & --
        & $4\times10^{-4}$
        \\

        Optimizer
        & AdamW
        & AdamW
        & AdamW
        & AdamW
        & AdamW
        \\

        LR Schedule
        & Cosine
        & Cosine
        & Cosine
        & Cosine
        & Cosine
        \\

        Warmup Ratio
        & 0.05
        & 0.05
        & 0.05
        & 0.05
        & 0.05
        \\

        Weight Decay
        & 0.05
        & 0.05
        & 0.05
        & 0.05
        & 0.05
        \\

        Max Gradient Norm
        & 1.0
        & 1.0
        & 1.0
        & 1.0
        & 1.0
        \\

        Max Sequence Length
        & 20,000
        & 20,000
        & 20,000
        & 20,000
        & 20,000
        \\

        Trainng NVIDIA GPUs(8)
        & H100(80GB)
        & H100(80GB)
        & A40(48GB)
        & H100(80GB)
        & A40(48GB)
        \\

        \bottomrule
    \end{tabular}
    }
\end{table}

\subsection{Evaluation Protocol}
\label{sec:supp_evaluation_protocol}

In the main paper, we follow the protocol adopted by RexOmni~\citep{jiang2026detect} and use \textbf{Recall}, \textbf{Precision}, and \textbf{F1 score} as the primary evaluation metrics at an IoU threshold of 0.5. 
We retain the top 100 predictions for each predicted category in each image. 
For each category, predictions from all evaluation images are then aggregated and matched against the corresponding ground-truth boxes to compute Recall and Precision, from which the F1 score is obtained as their harmonic mean. 
The final Recall, Precision, and F1 are averaged across categories. 
For DOTA2 \& SODA-A, we adopt the same category-wise evaluation procedure, but follow the standard DOTA2~\citep{xia2018dota} setting and retain up to 1,000 predictions per category for each image. 
For RSDense, however, the severe category imbalance makes category-averaged evaluation less representative of performance in extremely dense scenes. 
We therefore evaluate all predicted boxes and perform class-agnostic matching within each image. 
Recall, Precision, and F1 score are computed independently for each image and then averaged over all images in RSDense. 
The evaluation prompt contains only the ground-truth categories present in each image. 

Capped evaluation alone does not fully characterize the generation behavior of autoregressive detectors. 
For conventional detectors, predictions beyond the top-$K$ limit are typically removed only during evaluation, whereas every prediction generated by an autoregressive detector has already consumed decoding steps and computational budget. 
This distinction is particularly important in dense scenes, where a model may enter a box repetition loop and continue producing large numbers of redundant predictions after most objects have already been covered. 
Under top-$K$ evaluation, many of these late predictions can be excluded from the final metrics, obscuring the actual redundancy and termination behavior of the model. 
The truncation may also alter the observed Recall--Precision trade-off: valid detections beyond the cap can be discarded, limiting Recall, while redundant late predictions can be filtered out at the same time, potentially yielding an overly optimistic Precision relative to the model's complete output. 
Therefore, although capped evaluation is useful for maintaining comparability with existing benchmarks, it provides only a partial view of long-horizon autoregressive detection, where the ability to terminate generation after sufficiently covering the scene is itself an important aspect of both detection quality and inference efficiency. 

To complement the standard protocol, we further report an \textbf{all-prediction evaluation} in the supplementary material, where all predictions generated by the model are retained without imposing an additional per-image prediction cap. 
We otherwise follow the same category-wise evaluation procedure: predictions of each category are aggregated across all evaluation images to compute Recall and Precision, F1 is obtained as their harmonic mean, and the final metrics are averaged across categories. 
This protocol is not intended to replace the standard evaluation, but rather to provide a more complete characterization of \textbf{object discovery}, \textbf{prediction redundancy}, and \textbf{generation termination behavior}.
This evaluation is particularly relevant to WinAR, whose objective is not only to improve recall in dense scenes, but also to suppress the box repetition loop and allocate the finite decoding budget more effectively to discovering new objects.

\subsection{More Experiments}
\label{sec:more_experiments}

\begin{table*}[t]
    \centering
    \caption{Performance comparison between RexOmni-FHAR$^*$ and RexOmni-WinAR on COCO, DOTA2 \& SODA-A, RSDense, and SKU110K under the all-prediction evaluation setting. RexOmni-FHAR$^*$ denotes the RexOmni-FHAR baseline trained with long-sample oversampling.}
    \label{tab:all_box_detection_results}

    \setlength{\tabcolsep}{15pt}
    \renewcommand{\arraystretch}{1.15}
    \small

    \resizebox{\textwidth}{!}{%
        \begin{tabular}{llccc}
            \toprule
            Dataset & Method
            & Precision@0.50$\uparrow$
            & Recall@0.50$\uparrow$
            & F1@0.50$\uparrow$ \\
            \midrule

            COCO
            & FHAR*  & 63.28 & 67.84 & 65.48 \\
            & WinAR
            & 64.31 {\scriptsize $(+1.03)$}
            & 67.87 {\scriptsize $(+0.03)$}
            & 66.04 {\scriptsize $(+0.56)$} \\

            \midrule

            DOTA2 \& SODA-A
            & FHAR*  & 62.08 & 66.94 & 64.42 \\
            & WinAR
            & 62.79 {\scriptsize $(+0.71)$}
            & 68.40 {\scriptsize $(+1.46)$}
            & 65.17 {\scriptsize $(+0.75)$} \\

            \midrule

            SKU110K
            & FHAR*  & 53.89 & 81.75 & 64.96 \\
            & WinAR
            & 54.93 {\scriptsize $(+1.04)$}
            & 80.94 {\scriptsize $(-0.81)$}
            & 65.44 {\scriptsize $(+0.48)$} \\

            \midrule

            RSDense
            & FHAR*  & 55.90 & 38.95 & 45.91 \\
            & WinAR
            & 53.36 {\scriptsize $(-2.54)$}
            & 47.69 {\scriptsize $(+8.74)$}
            & 50.37 {\scriptsize $(+4.46)$} \\

            \bottomrule
        \end{tabular}%
    }
\end{table*}

\textbf{Additional Results on More Datasets.} 
Table~\ref{tab:all_box_detection_results} reports the results under the all-prediction evaluation setting, where every generated prediction is included without imposing a per-image prediction cap. WinAR consistently improves F1 over the exposure-matched FHAR$^*$ baseline across all evaluated datasets, while the gains become substantially larger as scene density increases. On COCO, WinAR preserves performance in ordinary scenes, improving F1 from 65.48 to 66.04. Similar improvements are observed on DOTA2 \& SODA-A and SKU110K, where F1 increases by 0.75 and 0.48 points, respectively. The advantage is most pronounced on RSDense: WinAR improves Recall from 38.95 to 47.69 $(+8.74)$ and F1 from 45.91 to 50.37 $(+4.46)$, despite a moderate decrease in Precision. These results further show that the benefit of WinAR is not an artifact of prediction truncation during evaluation. Instead, by suppressing redundant long-horizon generation, WinAR enables the autoregressive detector to allocate more of its decoding budget to previously uncovered objects, yielding increasingly pronounced gains in dense scenes while maintaining performance under ordinary scene densities. 

\begin{table}[t]
    \centering
    \caption{
        Comparison of inference throughput and training peak GPU memory.
        \textbf{Boxes/s} denotes the average number of predicted boxes per image per second,
        while \textbf{Valid Boxes/s} measures the throughput after removing redundant predictions. 
        Inference speed is measured on eight NVIDIA H100 GPUs.
    }
    \label{tab:efficiency_memory}

    \setlength{\tabcolsep}{13pt}
    \renewcommand{\arraystretch}{1.15}
    \small

    \resizebox{\columnwidth}{!}{
    \begin{tabular}{llccc}
        \toprule
        Method
        & Mode
        & Boxes/s$\uparrow$
        & Valid Boxes/s$\uparrow$
        & \makecell{Training Peak\\
                    GPU Memory (GB)$\downarrow$}
        \\
        \midrule

        RexOmni--FHAR$^*$
        & NTP
        & 6.05
        & 3.57
        & 57.62
        \\

        \rowcolor{blue!6}
        RexOmni-WinAR
        & NTP
        & \textbf{6.68}
        & \textbf{5.40}
        & \textbf{42.97}
        \\

        \midrule

        Locate Anything--FHAR$^*$
        & NTP
        & \textbf{5.46}
        & 5.14
        & 67.38
        \\

        \rowcolor{blue!6}
        Locate Anything-WinAR
        & NTP
        & 5.40
        & \textbf{5.37}
        & \textbf{44.43}
        \\

        \cmidrule(lr){1-5}

        Locate Anything-FHAR$^*$
        & PBD
        & \textbf{29.03}
        & 24.63
        & 67.38
        \\

        \rowcolor{blue!6}
        Locate Anything-WinAR
        & PBD
        & 26.15
        & \textbf{25.61}
        & \textbf{44.43}
        \\

        \bottomrule
    \end{tabular}
    }
\end{table}

\textbf{Training and Inference Efficiency.}
Table~\ref{tab:efficiency_memory} evaluates the training memory and inference efficiency of WinAR. 
WinAR substantially reduces peak GPU memory during training.
On RexOmni, peak memory decreases from 57.62~GB with FHAR$^*$ to 42.97~GB with WinAR, corresponding to a 25.4\% reduction. 
The reduction is even larger on LocateAnything, from 67.38~GB to 44.43~GB (34.1\%). 
This benefit stems from the window-wise training formulation: once a completed answer window is compressed, its raw token-level representations and associated activations can be discarded, substantially reducing the per-GPU memory requirement and allowing WinAR to fit within the 48~GB memory budget of NVIDIA A40 GPUs. 
At inference time, WinAR does not consistently improve raw prediction throughput because history compression and KV-cache prefilling introduce additional overhead at window boundaries. 
Nevertheless, WinAR consistently improves effective throughput by suppressing redundant predictions. 
On RexOmni, Valid Boxes/s increases from 3.57 to 5.40. 
For LocateAnything, Valid Boxes/s improves from 5.14 to 5.37 under NTP and from 24.63 to 25.61 under PBD, despite slight decreases in raw Boxes/s. 
Overall, WinAR substantially improves training scalability while allocating a larger fraction of the decoding budget to non-redundant predictions during inference. 

\textbf{Qualitative Comparison. }
The visualization results further demonstrate the effectiveness of WinAR in long-horizon dense grounding. Compared with FHAR$^*$, WinAR produces substantially fewer repeated predictions and recovers more distinct objects in crowded scenes. 
For polygon detection, WinAR also yields more complete and spatially consistent predictions, indicating that its benefit extends beyond axis-aligned boxes to more general geometric outputs.

\begin{figure}[h]
    \centering
    \includegraphics[width=\linewidth]{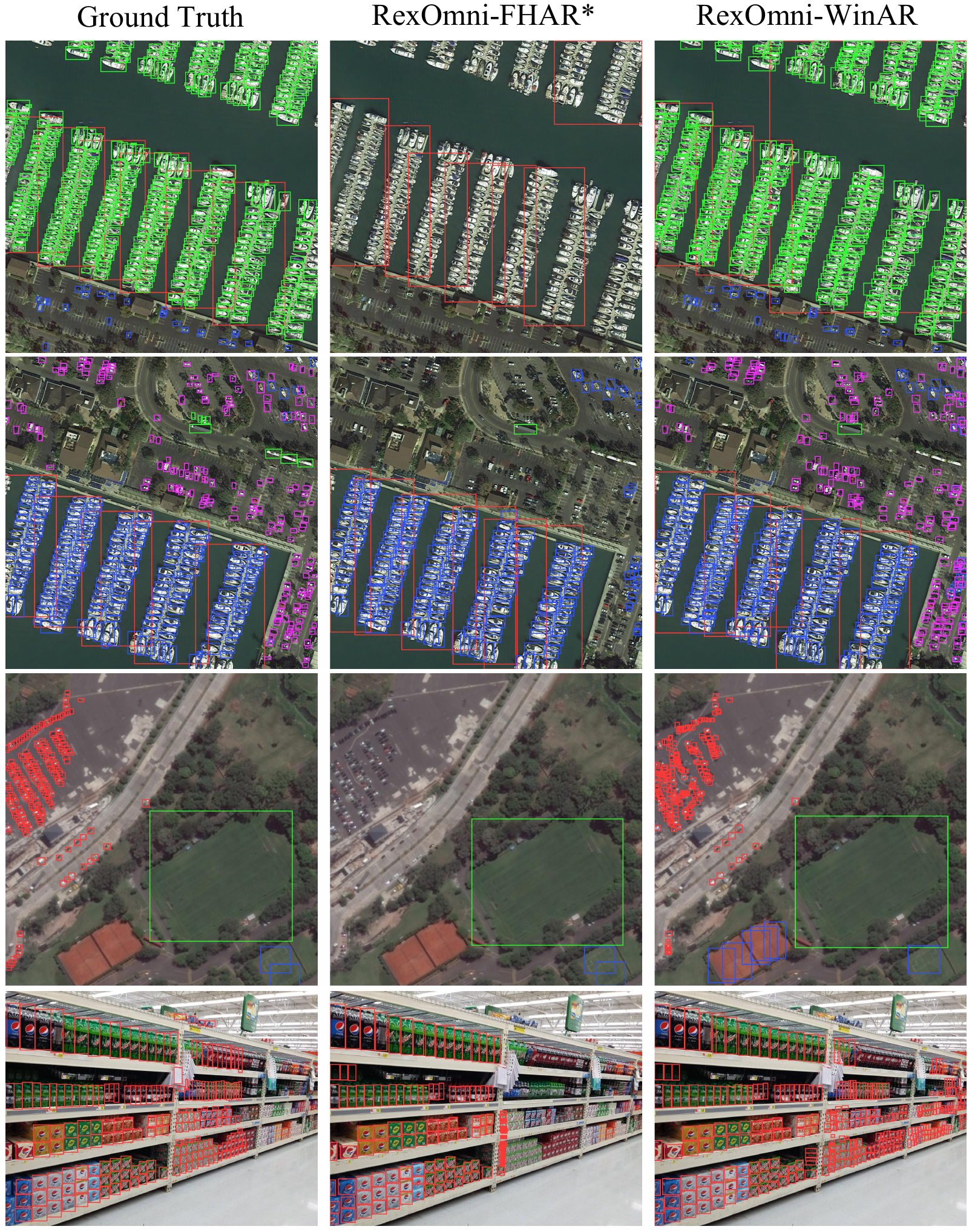}
    \caption{
        Visualization Comparison in Dense Scenes.
    }
    \label{fig5}
 \end{figure}

 \begin{figure}[t]
    \centering
    \includegraphics[width=\linewidth]{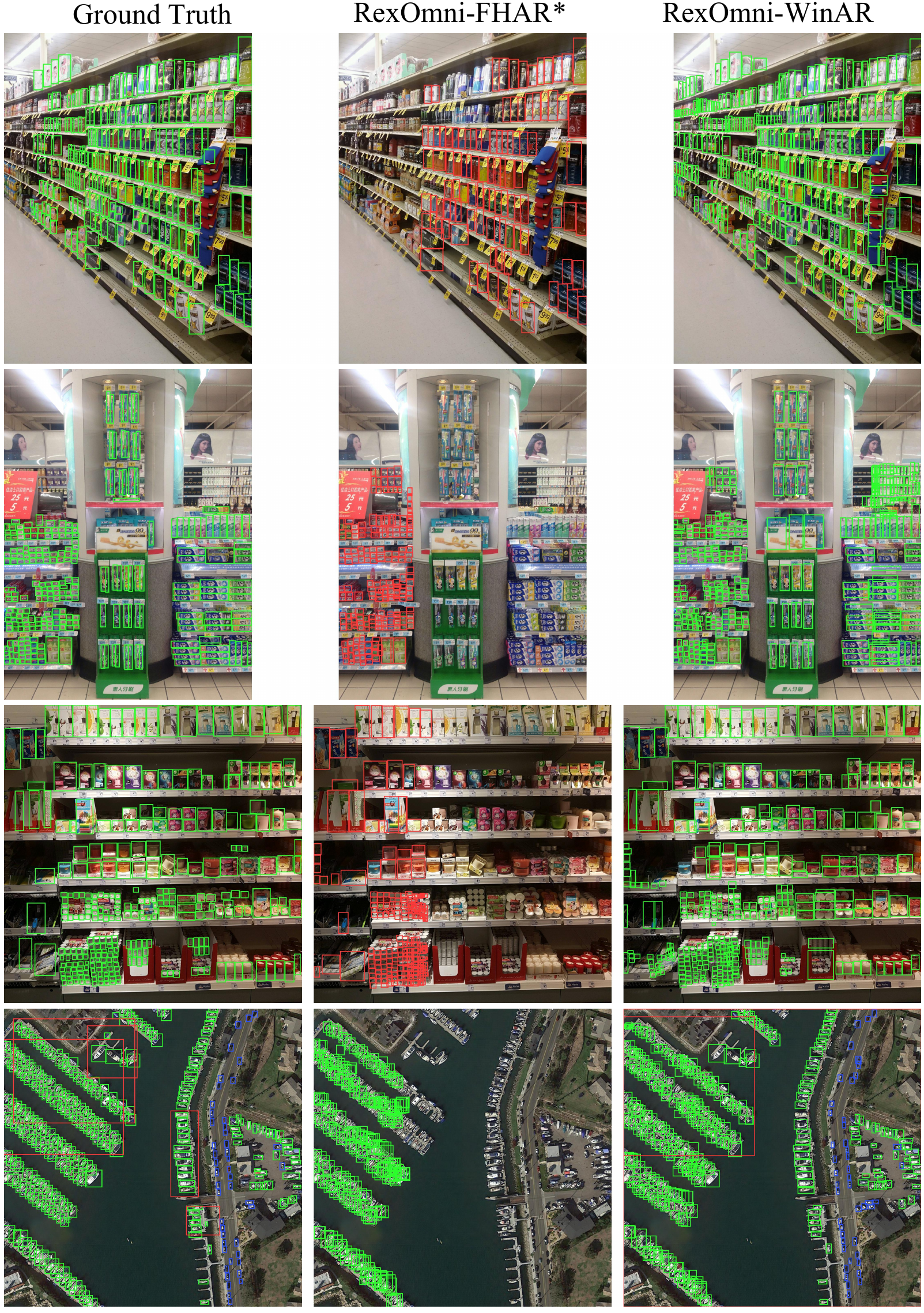}
    \caption{
        Visualization Comparison in Dense Scenes.
    }
    \label{fig6}
 \end{figure}

 \begin{figure}[t]
    \centering
    \includegraphics[width=\linewidth]{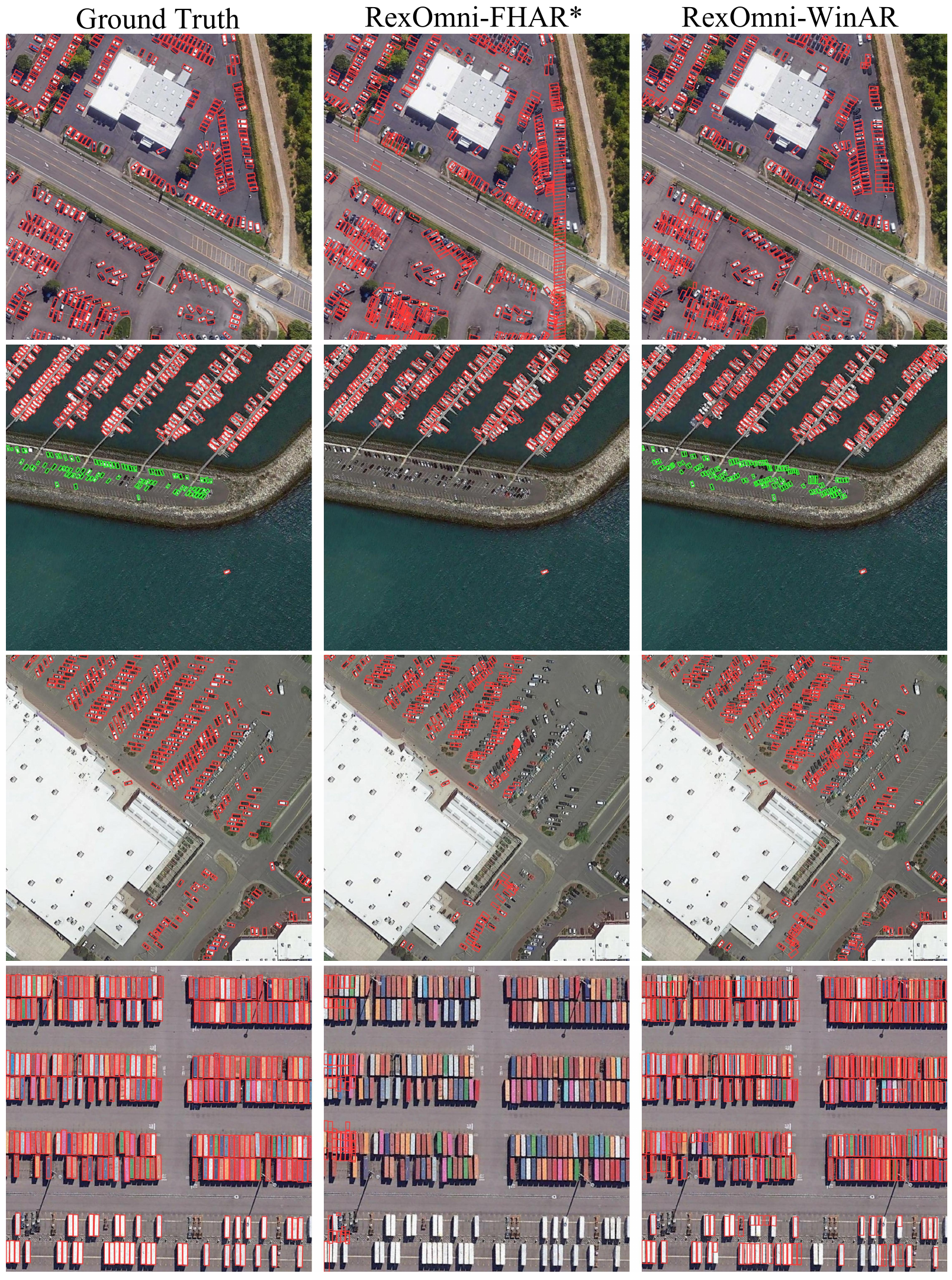}
    \caption{
        Visualization Comparison in Polygon Detection.
    }
    \label{fig7}
 \end{figure}

 \begin{figure}[t]
    \centering
    \includegraphics[width=\linewidth]{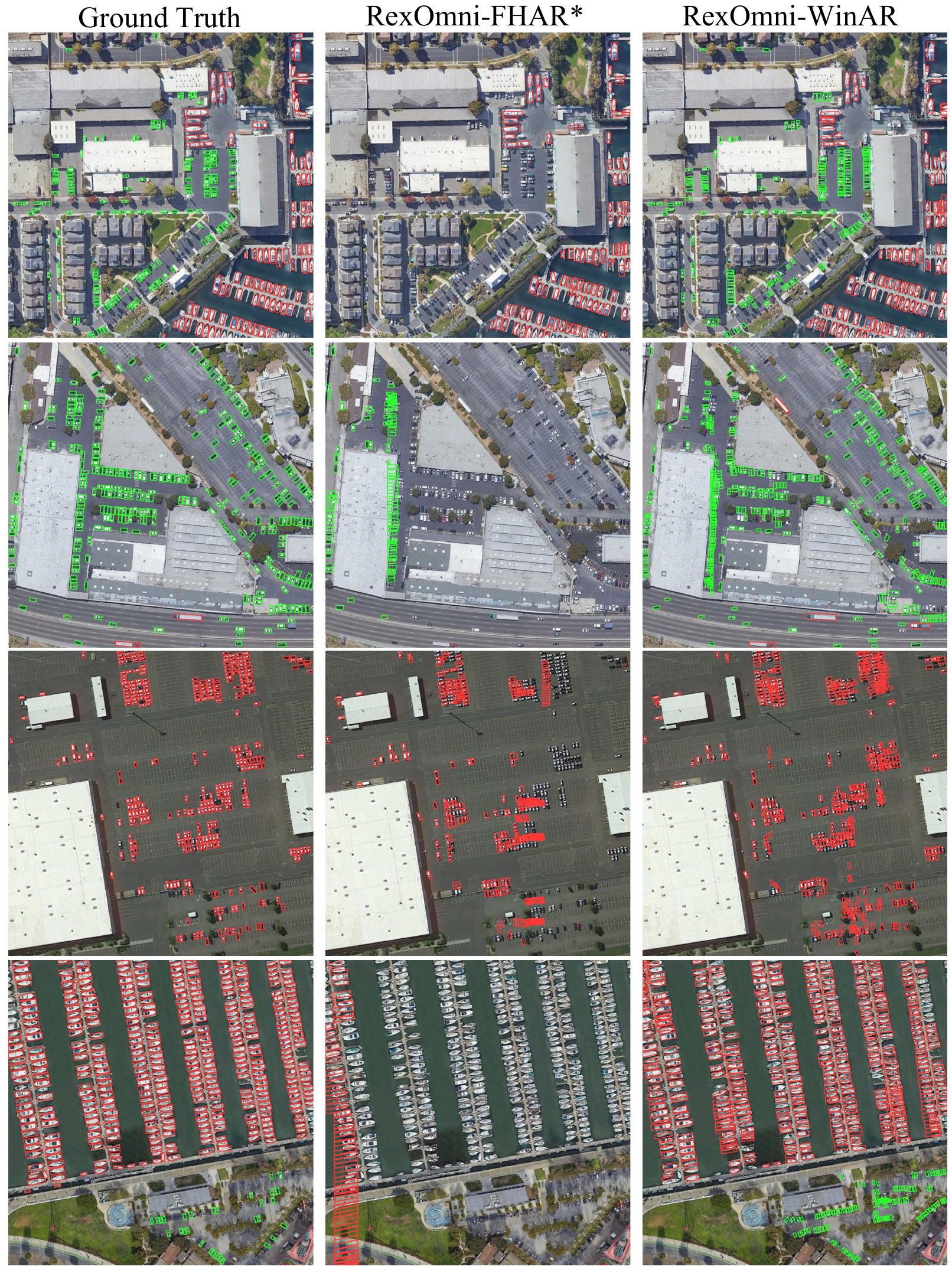}
    \caption{
        Visualization Comparison in Polygon Detection.
    }
    \label{fig8}
 \end{figure}

\end{document}